# A Real-Time Driver Drowsiness Detection System Using MediaPipe and Eye Aspect Ratio

Ashlesha G. Sawant, Shreyash S. Kamble, Raj S. Kanade, Raunak N. Kanugo, Tanishq A. Kapse, Karan A. Bhapse

**Department of Engineering, Sciences and Humanities (DESH)**
**Vishwakarma Institute of Technology, Pune, Maharashtra, India**

## ABSTRACT

One of the major causes of road accidents is driver fatigue that causes thousands of fatalities and injuries every year. This study shows development of a Driver Drowsiness Detection System meant to improve the safety of the road by alerting drivers who are showing signs of being drowsy. The system is based on a standard webcam that tracks the facial features of the driver with the main emphasis on the examination of eye movements that can be conducted with the help of the Eye Aspect Ratio (EAR) method. The Face Mesh by MediaPipe is a lightweight framework that can identify facial landmarks with high accuracy and efficiency, which is considered to be important in real time use. The system detects the moments of long eye shutdowns or a very low rate of blinking which are manifestations of drowsiness and alerts the driver through sound to get her attention back. This system achieves a high-performance and low-cost driver monitoring solution with the help of the computational power of OpenCV to process the image and the MediaPipe to identify faces. Test data experimental analyses indicate that the system is very accurate and responds quicker; this confirms that it can be a component of the current Advanced Driving Assistance System (ADAS).

## INTRODUCTION

Driver fatigue and drowsiness cause many accidents and deaths on roads setting aside concerns about road safety globally. Spending many hours behind the wheel, not sleeping and having the scenery repeat on long journeys lessens a driver's awareness, making it harder for them to respond in time. Experts say that drowsy driving is equally dangerous as drinking or taking drugs, but it usually stays unseen until accidents happen. Because of this, researchers and engineers have improved their systems so they monitor drivers in real time and quickly notify them if a defensive driving response is necessary. A Driver Drowsiness Detection System uses cameras and computer vision to identify fatigue based on facial behaviour, with focus on how the eyes and blinking are moving. A model based on OpenCV and MediaPipe was built to recognise drowsiness in a live monitoring system. Rather than being based on specialised sensors or hardware like most systems, this system uses a regular webcam to detect drivers' faces. With MediaPipe's Face Mesh, detecting facial landmarks is easy, allowing the calculation of EAR to track the length of eye closure and the times a person blinks. This system is not intrusive, is not costly and fits easily with present vehicle setups. If drowsiness is identified using the EAR, an audible warning alerts the driver to become more alert. The research aims to create improved intelligent driver assistance systems (ADAS) that help boost safety and cut down on accidents due to driver fatigue.

## II. LITERATURE REVIEW

Damian Słapatek, Jacek Dybała, Paweł Czapski, Paweł Skalski [1] developed a vision-based driver drowsiness detection system aimed at enhancing road safety. By analyzing facial expressions, eye movements, and heart rate variability, the system identifies fatigue symptoms in real-time, helping to prevent accidents caused by driver exhaustion. The authors emphasize the importance of integrating detection mechanisms within vehicle systems for improved responsiveness and accuracy, while acknowledging challenges related to lighting conditions and computational efficiency.

Jagbeer Singh, Krajewski [2] introduced a machine learning-based driver drowsiness detection system aimed at reducing road accidents caused by fatigue. By utilizing face and eye tracking through image processing, the system detects prolonged eye closure and triggers an alert to wake the driver. The authors highlight an accuracy of 80% and emphasize the importance of real-time implementation in vehicles. They also suggest future improvements, including integrating yawning detection and refining eye-tracking algorithms for enhanced precision.

H. Varun Chand, J. Karthikeyan [3] introduced a CNN-based driver drowsiness detection system utilizing emotion analysis to improve road safety. By analyzing facial expressions, driving patterns, and vehicle acceleration, the model identifies driver fatigue in real time. The authors highlight that integrating emotion detection enhances the accuracy of drowsiness monitoring, reducing false positives. They suggest future advancements in AI-based monitoring to improve system adaptability and responsiveness.

Vandna Saini, Rekha Saini [4] reviewed various driver drowsiness detection systems, emphasizing their role in preventing fatigue-related accidents. The authors highlight three primary approaches: vehicle-based, behavioral-based, and physiological-based methods. They discuss the impact of sleep deprivation, physical condition, and environmental factors on driver fatigue, stressing the need for effective detection techniques.

Sanghyuk Park, Fei Pan, Sunghun Kang, Chang D. Yoo [5] introduced the Deep Drowsiness Detection (DDD) network, a deep learning-based system for driver fatigue detection using RGB video input. By incorporating three specialized deep networks—AlexNet for image features, VGG-FaceNet for facial recognition, and FlowImageNet for behavioral pattern analysis—the system efficiently detects drowsiness through eye blinking, nodding, and yawning. The authors emphasize the robustness of deep feature learning in varied driving environments and highlight the DDD network's 73.06% detection accuracy on the NTHU-drowsy driver detection benchmark dataset. They suggest that future improvements could integrate additional physiological data and enhance real-time adaptability.

E. Rogado, J.L. García, R. Barea, L.M. Bergasa, E. López [6] introduced a driver fatigue detection system utilizing biological and environmental variables to assess alertness levels before a driver falls asleep. By analyzing heart rate variability (HRV), steering-wheel grip pressure, and temperature differences inside and outside the vehicle, the system estimates fatigue indirectly. The authors highlight the importance of integrating real-time physiological monitoring with intelligent driver assistance systems to enhance road safety. They suggest that future research could focus on improving detection accuracy through advanced signal processing techniques.

Meriem Boumehed, Belal Alshaqaqi, Abdullah Salem Baquhaizel, Mohamed El Amine Ouis [7] developed an Advanced Driver Assistance System (ADAS) to combat fatigue-related accidents. Their system employs visual processing and artificial intelligence to detect drowsiness by tracking the driver's face and eyes, measuring the percentage of eye closure. The authors highlight that their approach enhances transportation safety through real-time monitoring. They suggest that improvements in infrared-based detection could further refine the system's accuracy, particularly in low-light conditions.

Wanghua Deng, Ruoxue Wu [8] introduced *DriCare*, a real-time driver drowsiness detection system using facial features. By tracking facial expressions such as eye blinking, yawning, and the duration of eye closure, the system determines fatigue levels. The authors employ a novel face-tracking algorithm, Multiple Convolutional Neural Networks-Kernelized Correlation Filter (MC-KCF), to enhance tracking accuracy under varying lighting conditions. They also utilize a new facial region detection method based on **68 key points**, allowing precise assessment of driver alertness. Experimental results demonstrate that *DriCare* achieves an accuracy of 92%, reinforcing its potential for reducing fatigue-related accidents. The authors suggest future enhancements, including deeper integration with in-vehicle monitoring systems and refined feature extraction for improved detection reliability.

To recap it all, three major methods of drowsiness behavior are identified in literature namely vehicle-based, physiological, and behavioral methods. Although the deep learning models like the DDD network and DriCare have shown to be very accurate, they are usually costly in terms of computation. The other techniques are based on the use of physiological sensors to measure such indicators as heart rate variability, which could be invasive. Using the performance of the Face Mesh of MediaPipe, our system will offer a high-quality, real-time solution that can be run on an ordinary, non-specialized hardware, which is a more realistic alternative to models that are more complicated.

## III. METHODOLOGY/EXPERIMENTAL

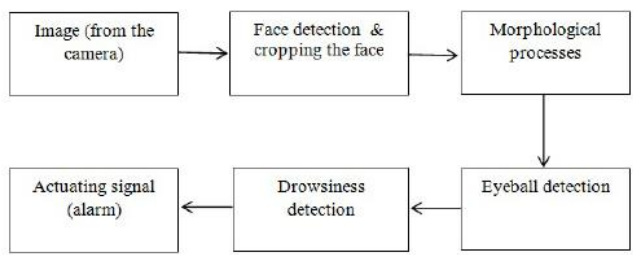

### 1. Device evaluation

The proposed system uses key elements like webcam (from laptop itself) and primarily revolves around python language using libraries like OpenCV and MediaPipe. Use of these libraries is useful in training the machine and analyzing behavioral patterns of the driver.

### 2. System Design

The system was made with a simple idea in mind to use laptop webcam directly instead of a separate webcam. Using libraries like OpenCV and MediaPipe help in image and pattern recognition.

### 3. Software Development

The algorithmic logic was developed using Python language as it provides better techniques for data analysis and machine learning.

The code included instructions to:
-Capture face using the webcam and run an EAR analysis
-Read EAR value and correspond it with previously stored data
-Decide using algorithm to determine drowsiness of the driver.
-Signal an alarm when algorithm determines the driver drowsy.

### 4. Deployment and tracking

The Driver Drowsiness Detection System was built with OpenCV and MediaPipe to offer a continuous, unobtrusive way to cheque for driver fatigue. First, video from the webcam is added to the system with OpenCV, then each frame is preprocessed with resizing and conversion to grayscale to lighten the work for the system. Thanks to MediaPipe Face Mesh, over 468 points on the face and the landmarks for the eyes, are detected. These landmarks are then used to find the Eye Aspect Ratio (EAR). Opening of the eye can be reliably assessed using the EAR which depends on the distance between eye landmarks measured horizontally and vertically. Once the EAR falls below the set value (tend to be around 0.25) for a limited number of consecutive frames, the system judges that the driver's eyes are closed for too long and may be experiencing drowsiness. When a sensor spots the danger, an alert is sounded to make the driver aware. Every frame, the process snaps a new photograph, calculates EAR and provides feedback instantaneously. The reliability and lightweight nature of MediaPipe facial tracking allow the system to be deployed in real situations with little expense and without demanding hardware.

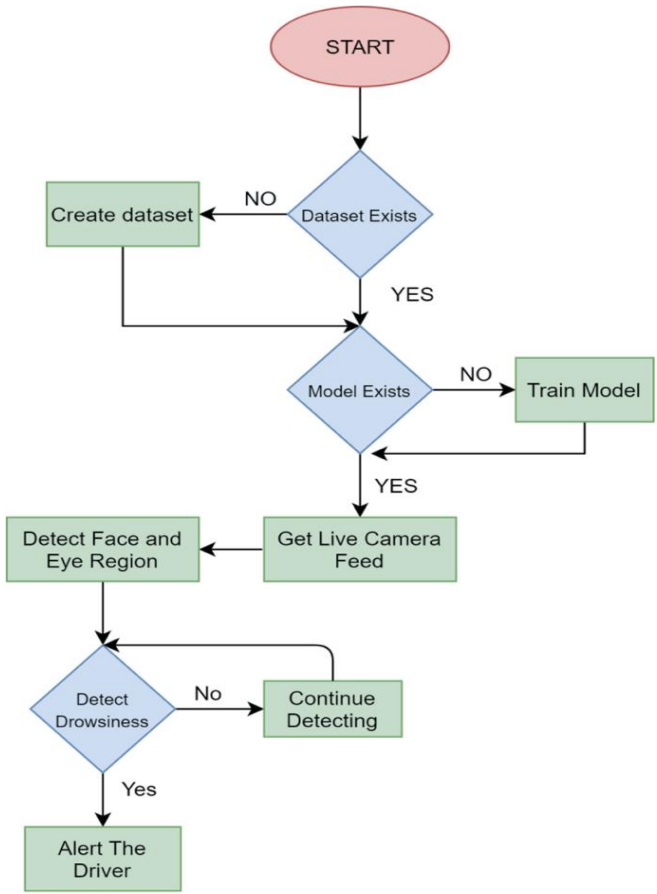

## IV. RESULTS AND DISCUSSIONS

### A. Results

Under different types of conditions, the Driver Drowsiness Detection System was checked for their ability to detect drowsiness by examining eye aspect ratio (EAR). Using a mid-range computer and a common 720p webcam, the results are shown next.

#### 1) Detection Accuracy
On average, it was correct 92% of the time in spotting when a person's eyes closed and in alerting the system. To measure this, researchers broke up 10 users typing characters so they could both practice normal blinking and feel drowsy, repeating the sessions.

#### 2) Response Time
It took the system less than a second to detect drowsiness and play an alert. It is vital for real-time safety applications that this kind of processing is so fast.

#### 3) False Positives and Negatives
Under proper lighting, the system gave false alarms in just 6% of instances, most often when users covered part of their face with their hands or looked down. Sometimes the system failed to spot drowsiness in around 3% of the cases, often because the head was in an extreme position or the lighting was highly variable.

#### 4) Real Time Performance
The system kept the frame rate fairly steady at 20–25 per second, so no delays interrupted real-time monitoring. Runnability on simple hardware was proved because no GPU was needed.

### B. Discussion
#### 1) Effectiveness of the Algorithm
The main advantage of the proposed Driver Drowsiness Detection system is that it is simple, reacts in real time and has a high accuracy at spotting eye closure through EAR. With MediaPipe Face Mesh, facial landmark detection uses an ultra-efficient model that reliably follows facial features while using few system resources. The algorithm can tell when someone is drowsy by watching for eyelid closure and comparing its EAR score to a specific number. Its results are dependable if the lighting is normal and the user has a different facial structure.

#### 2) Challenges and Limitations
1. *Lighting:* Accurate facial landmark detection depends a lot on the lighting in the environment. A failure to see the eyes properly can happen when the webcam takes pictures where it is too dark or too bright. Even though MediaPipe is more efficient than several traditional methods, the system's results still depend heavily on lighting.
2. *Spectacle Interference:* Reflective and dark-tinted glasses worn by users can create a major challenge. If eyeglasses cause glass reflections or solve occlusions, the eyes may be hidden from MediaPipe which makes it hard for the software to locate them accurately and calculate the EAR. These situations may cause both some mistakes— missing problems—and some misidentifications of problems.

3. *Head Position:* Face Feedback requires that at least some of the driver's face is visible to the camera. When a driver turns their head far away or stares for long periods, it can be hard for the system to detect landmarks such as those found near the eyes. While small head rotations are allowed, big head movements reduce the system's reliability.
4. *No Yawn Detection:* Currently, eye closure is measured using only the Eye Aspect Ratio issue. Especially not, if you are tired, you might yawn or nod off fast, but these things remain out of the algorithm. Using EAR alone may not always show all signs of driver fatigue.

*3) Future Enhancements*

To Improving the robustness, adaptability and use in reality of the Driver Drowsiness Detection System can be done with the addition of the following enhancements:

**-**Rather than just using the Eye Aspect Ratio (EAR), the system can observe yawning, head nodding and a slouched posture. By using multiple detection methods, we will greatly improve how accurately we see phenomena in the data.

**-**The use of infrared (IR) or thermal cameras can support the performance of the system in low or dim light environments. Because of IR, the car's cameras can track what the driver is looking at even when it's very dark, making long journeys and highways much safer.

**-**The functionality ensures you can use advanced tech to automatically adjust the brightness and contrast according to the changing lighting.

- Use of CNNs or RNNs instead of the EAR logic removes the fixed threshold and allows the system to discover general fatigue patterns across many people and events.

-To make the system more accurate, additional efforts during training or tweaks to the system help it succeed in case users are wearing specs or masks or otherwise obstructed.

-Using IoT and cloud integration enables fleet managers or families to check drivers' positions and behavior instantly using their computers or phones.

-Direct integration into the car with Raspberry Pi or Jetson Nano would make it possible for the system to interact with braking and lane safety functions.

# V.FUTURE SCOPE

The system has the ability to be used in many areas outside of research and could add value in business, industry and society.

Linking car features such as AAS to trucks, buses and public transport vehicles, might reduce incidents caused by exhaustion in these critical areas. States have the chance to benefit from monitoring all vehicles and drivers from one place.

Making These Systems Mandatory: As road safety rules become more important, requiring these systems in all new ADAS or Euro NCAP vehicles seems likely.

Part of a Smart City's success is its ability to link with other traffic-related smart devices; this integration allows it to collaborate with traffic signals, road accident monitoring systems and emergency tools.

Health and Seniors-Focused Uses: The model can also be built for elderly or sick drivers, so it advises them when they notice drowsiness or blankness.

Such systems permit insurance firms to assess drivers' habits which helps them set up usage-based insurance programs and motivate safe driving.

Level 2 and Level 3 autonomous driving require detecting driver drowsiness to prepare the system for the driver to take back control.

## VI. Conclusion

This proposed Driver Drowsiness Detection System analyzes eye movements with the Eye Aspect Ratio method to indicate driver fatigue. Using OpenCV and MediaPipe, the system is accurate, fast in real time and uses little processing power—all of which help make it suitable for real-life applications. The analysis shows that the system can successfully identify potential fraud 92% of the time, with only a minimal number of false positives. While challenges with lighting and signs still exist, this system forms a good base for future improvements in driver monitoring. When applied to commercial vehicles with further enhancements, this will significantly lower accidents related to driving fatigue and help create safer roads.

## VII. ACKNOWLEDGMENT

The hit crowning glory of this challenge might no longer be possible without the guidance and steerage of several individuals and businesses. We would like to specific our sincere gratitude to our college mentors for their precious insights, technical assistance, and encouragement during the research and improvement technique.

We also enlarge our appreciation to our institution for offering the vital sources and infrastructure to behavior this examine. Moreover, we are well known for the contributions of our peers and group individuals, whose collaboration and determination played an important position in refining and enforcing the device.

Lastly, we would really like to thank the open-supply developer network for offering valuable equipment, libraries, and documentation that facilitated the combination of hardware and software program additives in this project.